\documentclass[sigconf]{acmart}
\usepackage[T1]{fontenc}
\usepackage{aecompl}
\usepackage{booktabs} 
\usepackage{graphicx}
\usepackage{caption}
\usepackage{subfigure}
\usepackage{float}
\usepackage{mathrsfs}
\setcopyright{rightsretained}

\begin{document}
\title{Imagination Based Sample Construction for Zero-Shot Learning}

\copyrightyear{2018}
\acmYear{2018}
\setcopyright{acmcopyright}
\acmConference[SIGIR '18]{The 41st International ACM SIGIR Conference on Research and Development in Information Retrieval}{July 8--12, 2018}{Ann Arbor, MI, USA}
\acmBooktitle{SIGIR '18: The 41st International ACM SIGIR Conference on Research and Development in Information Retrieval, July 8--12, 2018, Ann Arbor, MI, USA}
\acmPrice{15.00}
\acmDOI{10.1145/3209978.3210096} \acmISBN{978-1-4503-5657-2/18/07}

\author{Gang Yang}
\affiliation{
  \institution{Renmin University of China}
  \streetaddress{No. 59 Zhongguancun Street, Haidian District}
  \city{Beijing}
  \state{China}
  \postcode{100872}
}
\email{yanggang@ruc.edu.cn}

\author{Jinlu Liu}
\affiliation{
  \institution{Renmin University of China}
  \streetaddress{No. 59 Zhongguancun Street, Haidian District}
  \city{Beijing}
  \state{China}
  \postcode{100872}
}
\email{liujinlu@ruc.edu.cn}

\author{Xirong Li}
\authornote{Corresponding author.}
\affiliation{
  \institution{Renmin University of China}
  \streetaddress{No. 59 Zhongguancun Street, Haidian District}
  \city{Beijing}
  \state{China}
  \postcode{100872}
}
\email{xirong@ruc.edu.cn}

\begin{abstract}

Zero-shot learning (ZSL) which aims to recognize unseen classes with no labeled training sample, efficiently tackles the problem of missing labeled data in image retrieval. Nowadays there are mainly two types of popular methods for ZSL to recognize images of unseen classes: probabilistic reasoning and feature projection. Different from these existing types of methods, we propose a new method: sample construction to deal with the problem of ZSL.
Our proposed method, called Imagination Based Sample Construction (IBSC), innovatively constructs image samples of target classes in feature space by mimicking human associative cognition process. Based on an association between attribute and feature, target samples are constructed from different parts of various samples. Furthermore, dissimilarity representation is employed to select high-quality constructed samples which are used as labeled data to train a specific classifier for those unseen classes. In this way, zero-shot learning is turned into a supervised learning problem. As far as we know, it is the first work to construct samples for ZSL thus, our work is viewed as a baseline for future sample construction methods. Experiments on four benchmark datasets show the superiority of our proposed method.
\end{abstract}



\maketitle

\section{Introduction}

Development of image retrieval, especially of fine-grained image retrieval is more or less impeded by the problem of missing labeled data due to increasing annotation costs. As zero-shot learning (ZSL) realizes image classification of certain classes which have no labeled training samples, it has been drawing much attention in recent years \cite{6571196}.
But task of ZSL is difficult because it lacks labeled training samples of some classes, called unseen classes, to directly train classifiers. Just given labeled training samples of seen classes, ZSL aims to achieve unseen classes recognition by building relationship between unseen classes and seen classes. 
Nowadays, there are mainly two popular methods for ZSL: probability reasoning based on attribute prediction \cite{jayaraman2014zero} and feature projection among different embeddings \cite{li2015zero}. The first method usually predicts attribute probability to calculate class maximum likelihood while the second method mainly bridges different embedding spaces to exploit space projection and feature mapping for ZSL. However, existing methods have several flaws. On the one hand, there is inherent error accumulation in probabilistic reasoning. On the other hand, most embedding methods apply complex deep network to realize space projection, which is widely believed to have little interpretability of projection process and takes a lot of time to train the network. Different from existing two types of methods, we propose a new type of method for ZSL: sample construction. Our proposed method, Imagination Based Sample Construction (IBSC), is based on human associative cognition process to directly construct samples of unseen classes. Thus, unseen class recognition can be realized via learning from the constructed samples. 

\begin{figure}
\includegraphics[height=1.49in, width=3.4in]{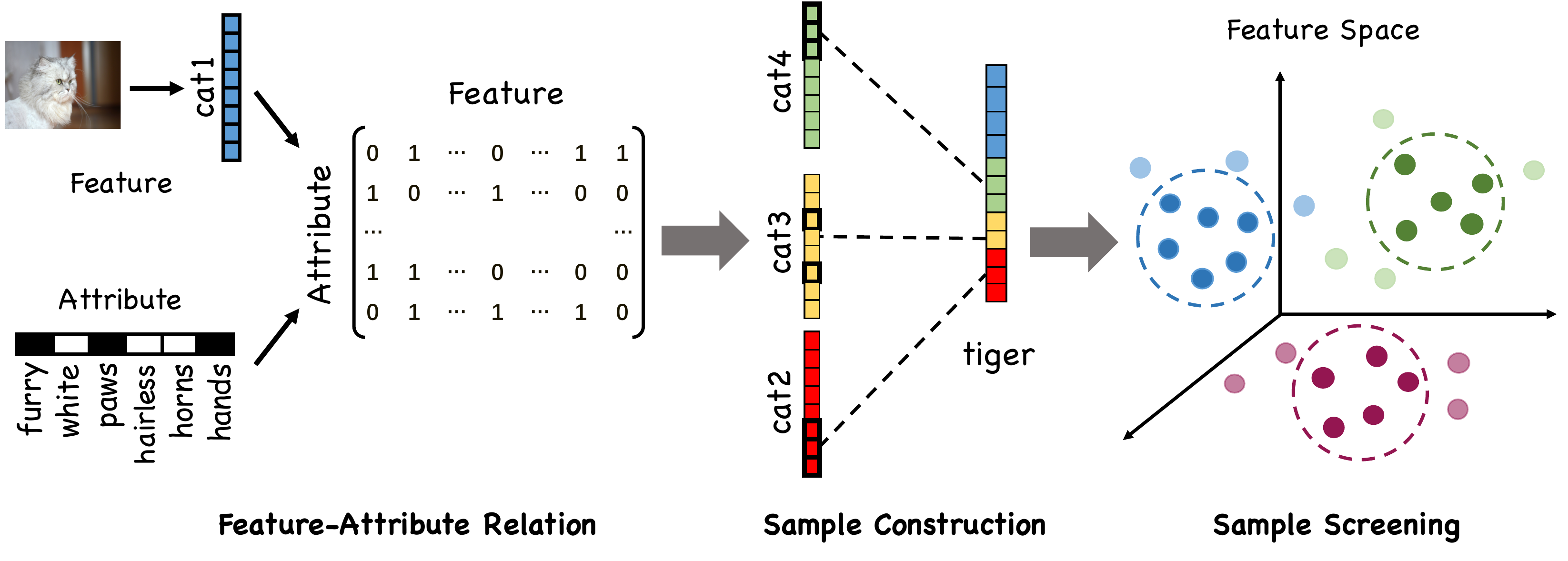}
\caption{Imagination Based Sample Construction: samples of unseen class ``tiger'' are constructed from samples of its similar class ``cat'' in feature space.}
\label{framework}
\end{figure}

Human can visualize unseen objects through referring some already known objects and assembling their visual components based on imagination \cite{Papadopoulos2011Noncategorical}. A human, who never see a tiger before but has seen some cats yet, can speculate the species at the first sight of a real tiger if he knows the description of tiger or attribute relationship between tiger and cat. By mimicking human associative cognition process, we construct samples of unseen classes from samples of seen classes in feature space, based on a relationship between feature and attribute.
Our proposed method is schematically displayed in Figure \ref{framework}. As shown in Figure \ref{framework}, each attribute is related to specific dimensions of image feature. For example, samples which don't have attribute ``paws'' are different from samples with ``paws'' in certain feature dimensions. Based on attribute-feature relation, an image feature can be reconstructed from other samples to express different attributes. If use feature dimension related to ``paws'' to replace original feature dimension of samples without ``paws'', the reconstructed samples will have a new characteristic. Generally, it is reasonable to choose seen classes with large similarity to unseen classes as reference basis when constructing target samples of unseen classes. After samples are constructed through splicing different samples of seen classes, the constructed samples of higher quality need to be picked out. Hence, we adopt the idea about \textit{dissimilarity representation} \cite{lee2012bridging} to measure representativeness of the constructed samples.
Although our method is designed to classify images of unseen classes, no more new classifiers need constructing, as ZSL has been simplified into a traditional supervised classification problem where most existing classifiers can be used.
We experiment on four benchmark datasets. Compared with state-of-the-art approaches, comprehensive results demonstrate the superiority of our proposed method. Furthermore, our work can be viewed as a baseline for future sample construction works for ZSL.

\section{Imagination Based Sample Construction}

We propose a new method, Imagination Based Sample Construction (IBSC), to directly construct exemplar samples of unseen classes. Problem definition is as follows. 
Classes are split into two types: seen classes $\mathcal{S}={\{S_i\}}^{K^s}_{i=1}$ and unseen classes $\mathcal{U}={\{U_i\}}^{K^u}_{i=1}$. Only labeled samples $\mathcal{D}={\{(x_i,y_i)\}}^n_{i=1}$ of seen classes are given. Attribute $\mathcal{A}={\{A_i\}}^K_{i=1}$ of each class is utilized to associate unseen classes with seen classes ($K=K^s+K^u$), where $A_i=(a^1_i,...,a^d_i)$ is a $d$-dimensional vector. Our target is to recognize unseen classes which have no labeled data at training stage. We firstly build a relationship between attribute and feature. This relationship indicates whether an image feature can characterize certain attributes that is to say, attribute-relevant feature is selected by the relationship. Then we choose similar (source) classes to construct samples of unseen (target) classes. When constructing new samples, attribute difference between source classes and target classes is measured to instruct the process of attribute-relevant feature substitution. After target samples are constructed, dissimilarity representation is adopted to filter out unsuitable constructed samples and to reserve the most representable samples which are used to train classifiers for unseen classes. Then ZSL problem is turned into a supervised learning problem.

\subsection{Feature Selection per Attribute}

Class attributes are related to partial image features strongly. For example, features related to attribute ``paws'' show what animals with paws look like in an image. That is to say, different combinations of feature dimension express diverse attribute characteristics. If relationship between attribute and feature is built, samples can be constructed from different features to characterize unseen class attributes. Thus, we aim to construct an attribute-feature relation as shown in the left part of Figure \ref{framework}. We use relation value $R_{ij}$ to denote relation between attribute and feature, where $R_{ij}=\{0,1\}$. And value $1$ means that the $j$-th dimension of feature can distinguish different classes in terms of the $i$-th dimension of attribute while value $0$ is opposite. 
For efficient feature selection, we employ linear support vector classification (SVC) with $\ell_1$ regularization to select feature per attribute. SVC can automatically select features relevant to certain attribute. The selected features have class distinguishing ability in terms of different attributes, which are called attribute-relevant features in this paper.

\subsection{Sample Construction}

Sample construction in our method refers to constructing new samples of unseen classes by splicing relevant parts in the selected samples of seen classes. Seen classes which contain these samples are called source classes and the selected samples are called source samples in this paper. To realize sample construction, we take three steps as follows.

\textbf{Step 1. Source Classes Selection} 
Because samples are constructed to train a classifier for unseen classes, the constructed samples should maximally characterize unseen classes. Samples constructed from source classes which have large similarity and small attribute difference with unseen classes should have better characteristic expression ability. Under this assumption, we use class similarity and attribute difference to instruct source class selection. The similarity between two classes is measured by $\ell_2$ distance of class attribute vectors: 
\begin{equation}
\label{classsim}
{\phi}_{ij}=\frac{{\Vert{A_i}-{A_j}\Vert}_2-\mu_1}{\sigma_1}
\end{equation}
where $\mu_1/\sigma_1$ is the mean value/standard deviation of $\ell_2$ distances between any two classes. Attribute difference is defined as Eq. (\ref{attrdif}):
\begin{equation}
\label{attrdif}
{\psi}_{ij}=\frac{\sum^{d}_{n=1}{|a^n_i-a^n_j|}-\mu_2}{\sigma_2}
\end{equation}
where $a^n_i/a^n_j$ is the $n$-th dimension of attribute $A_i/A_j$. Similarly, $\mu_2/\sigma_2$ is the mean value/standard deviation of total absolute difference of attribute values.

A source class set $\{S_1,...,S_k\}$ of unseen class $U_i$, is selected among seen classes by class similarity values. In source class set, source class $S_1$ has the largest class similarity to $U_i$ while $S_k$ has the smallest class similarity. Regarding class attributes, combination of source class attributes can be viewed as a virtual attribute vector of an unseen class. For examples, if a real attribute vector of unseen class $U_i$ is represented in binary value such as $A_i=(0,1,1,...,0,1,0)$, the corresponding virtual attribute vector should have small Hamming distance to $A_i$. Hamming distance is calculated as a reference to determine the number $k$ of source classes. We put more emphasis on selecting the most similar class $S_1$ because feature splicing is made on samples of the most similar class to construct new samples. When selecting more suitable seen classes as the most similar source classes of all unseen classes, we optimize:
\begin{equation}
\label{arg}
\mathop{\arg\min}_{\{S_j\}}\sum^{K^u}_{i=1}{{\phi}_{ij}+{\psi}_{ij}}
\end{equation}
$\{S_j\}$ is the most similar source classes set of unseen classes after adjustment by optimizing Eq. (\ref{arg}). After adjustment, the most similar classes of unseen classes have no repetitive classes, which enhance inter-class distinguishing degree when sample construction. That is to say, the second similar class $S_2$ will replace the original most similar class $S_1$ to be the new $S_1$ if total value of Eq. (\ref{arg}) is smaller when original $S_1$ is used as the most similar class of other unseen class. Adjusting in this way, inter-class discriminability and whole similarity between source classes and unseen classes are simultaneously optimized. 

\begin{figure}
\includegraphics[height=0.84in, width=2.9in]{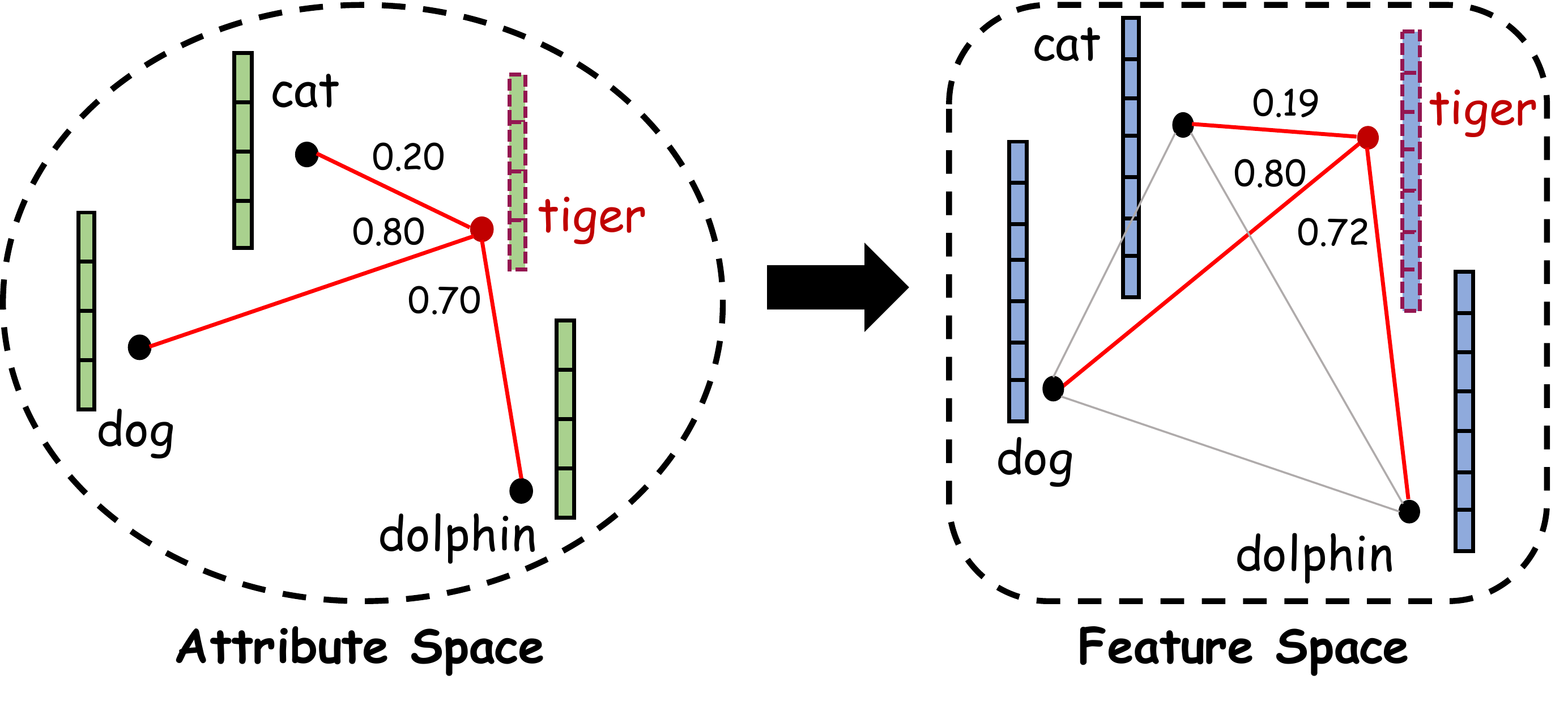}
\caption{\textit{Dissimilarity representation} bridges attribute space and feature space: a class can be represented by relative distance to other classes in different spaces. By knowing dissimilarity representation of unseen class ``tiger'' in attribute space is $(0.20,0.80,0.70)$, we can infer the class with similar dissimilarity representation $(0.19,0.80,0.72)$ in feature space to be ``tiger''.}
\label{diss_space}
\end{figure}

\textbf{Step 2. Sample Selection}
We use all samples of source class $S_1$ selected via above strategies to construct target samples of unseen class $U_i$. Features which are relevant to different attributes between $S_1$ and $U_i$ are replaced by new features. The new features are picked from samples of other classes in $\{S_2,...,S_k\}$. Thus, it is necessary to make reasonable selection of these samples in $\{S_2,...,S_k\}$.
Knowing which dimensions of feature are attribute-irrelevant, we define 
those dimensions where relation value $R$ equals to $0$ as environment information. To select more reasonable samples of other classes in $\{S_2,...,S_k\}$, we employ environment information to measure fitness degree of a sample. Fitness degree is measured by $\ell_2$ distance of environment information among different samples and samples with more similar environment information are selected to be candidate source samples when constructing target samples. Furthermore, we pre-train classifiers for each attribute by SVM which are used to measure attribute prediction capability by attribute prediction probability. Samples selected by environment information with the largest attribute prediction probability are most suitable to be used in sample construction.

\textbf{Step 3. Feature Construction}
Target samples of unseen classes can be constructed based on samples of source classes. Main process of construction is that features which are related to attributes shared by the most similar source class and unseen class are retained while the rest features are replaced by corresponding sample features of other classes. New features will be picked from samples selected at previous steps, which have strong attribute prediction ability and similar environment information with original source samples. Process of feature replacement is repeated until features of all attributes of unseen class are spliced on the source samples, as shown in the middle part of Figure \ref{framework}. Hence, the constructed features can be viewed as samples belonging to unseen classes, which generally characterize unseen classes.  

\subsection{Sample Screening}
There are some noisy samples in the constructed samples which are not suitable to  represent unseen classes. Because \textit{dissimilarity representation} can bridge different spaces to represent a class in an unified form, we apply dissimilarity representation to screen the constructed samples of higher quality \cite{lee2012bridging}. As shown in Figure \ref{diss_space}, classes in different spaces are represented in different forms but they are unified into one form which is expressed by the relative relation with other classes. That is, a well constructed ``tiger'' sample in feature space has the same dissimilarity representation of ``tiger'' in attribute space. So dissimilarity representation enables class attribute to instruct sample screening in feature space. Here we use Eq. (\ref{space}) to represent relative relation among classes:
\begin{equation}
\label{space}
d_{ij}=\frac{{\lVert{c_i-c_j}\rVert}^2_2}{\theta^2} 
\end{equation}

In attribute space, $c$ is a class attribute vector while in feature space, $c$ is the center of each class. In two spaces, $\theta^2$ has the same meaning: sum of $\ell_2$ distance among classes. Unseen classes dissimilarity representation in attribute space is denoted as a normalized vector $D_a=(d_1,...,d_{K^s})$. As aiming to screen the constructed samples of high-quality, we denote each sample instead of each class in form of normalized vector $D_f=(d_1,...,d_{K^s})$ in feature space, where relative relation is measured between each sample and centers of other classes. In terms of dissimilarity representation, quality of the constructed samples is measured by difference value $|D_f-D_a|$, which is to say that samples of small difference value are screened to be training data of unseen classes. Given abundant screened samples, zero-shot learning is turned into a supervised learning problem where unseen classes can be classified by existing classifiers.

\section{Experiments and Analysis}

\textbf{Datasets} We conduct experiments on four benchmark datasets: Animals with Attributes (AwA1) \cite{6571196}, Animals with Attributes 2 (AwA2) \cite{XianLSA17}, SUN Attribute (SUN) \cite{Hays2012SUN}, and Caltech-USCD-Birds-200-2011 (CUB) \cite{WahCUB_200_2011}. Both AwA1 and AwA2 have 40 seen classes and 10 unseen classes, totally containing 30,475 and 37,322 images. As for SUN, there are 707 seen classes and 10 unseen classes, adding up to 14,340 images. CUB is a set which has 200 bird species with 150/50 seen/unseen classes. Attribute dimensions of four datasets are 85, 85,102 and 312. 

\textbf{Setup} In experiments, continuous attribute is utilized to measure class similarity and attribute difference while binary attribute is used to determine the quantity $k$ of source classes of each unseen class. The quantity $k$ is set to 5 in our experiments. Since deep Convolutional Neural Network (CNN) feature has been proven to have the best feature expression ability of images, we use fc7 layer output of VGG-19 network as image feature which is a 4096-dimensional vector \cite{SimonyanZ14a}.

\begin{figure}
\subfigure[chimpanzee]{
\includegraphics[height=0.81in, width=1.42in]{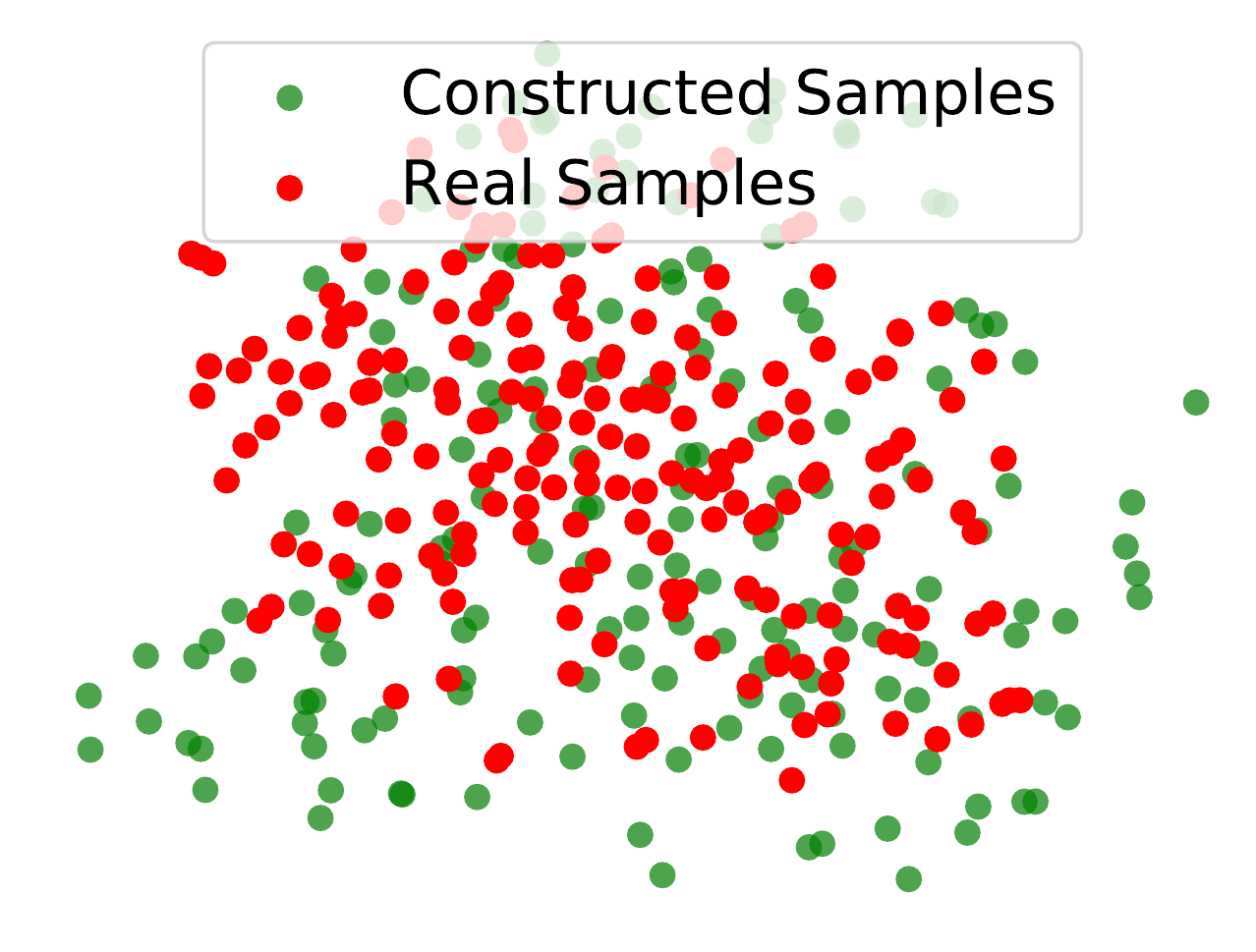}
}
\subfigure[humpback whale]{
\includegraphics[height=0.81in, width=1.42in]{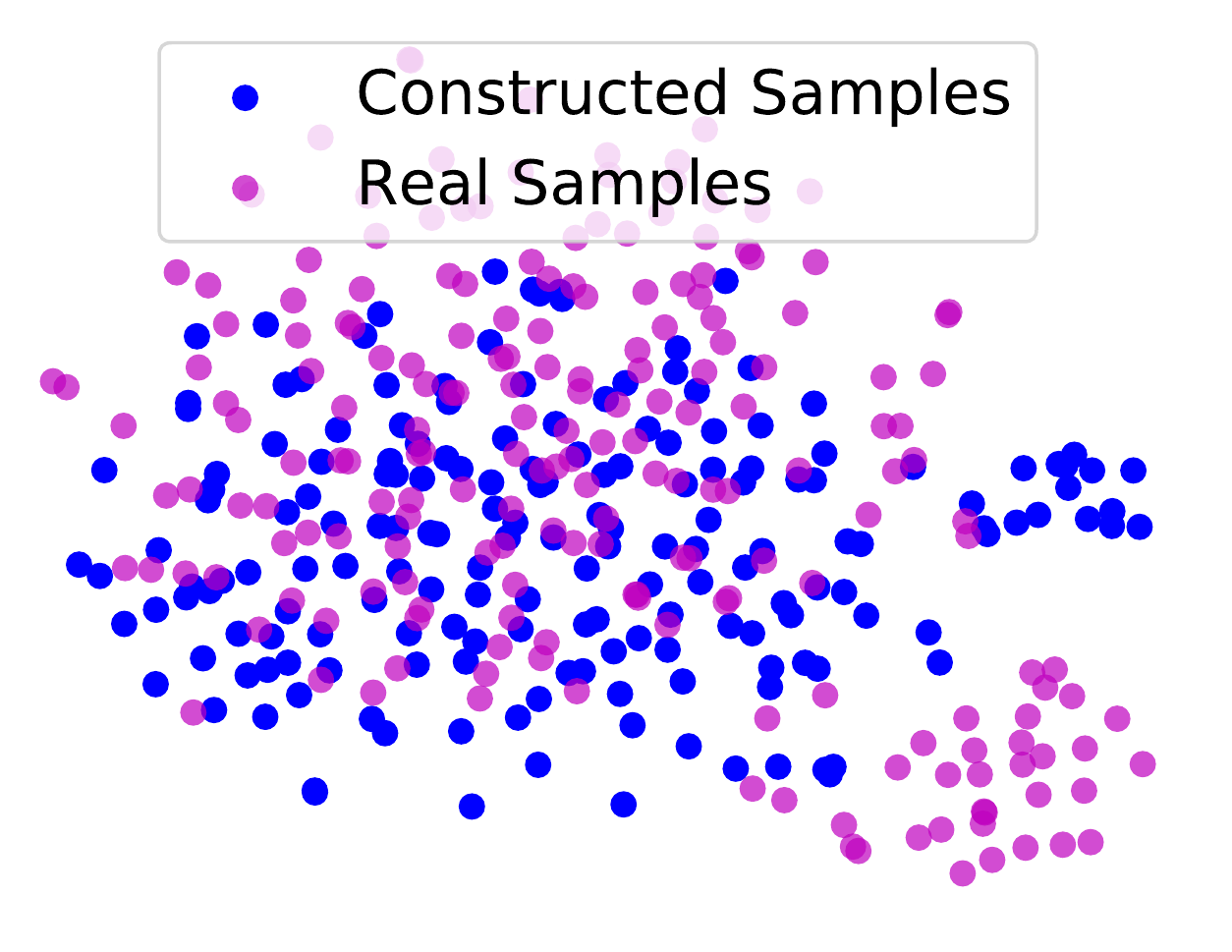}
}
\caption{t-SNE visualization of the constructed samples after screening and real samples of two unseen classes in AwA1 (better viewed in color).}
\label{tsne}
\end{figure}

\textbf{Analysis} As shown in Figure \ref{tsne}, the constructed samples distribute closely to real samples of of unseen classes. Thus, the constructed samples have strong ability to characterize unseen classes and using them as training data is proven to be reasonable. We compare top-1 accuracy with several state-of-the-art methods on four benchmark datasets. Comprehensive results comparison is shown in Table \ref{results}. It can be clearly seen that our method outperforms on three datasets, especially on AwA1 and SUN. Our method constructs representative samples of high quality as shown in Figure \ref{tsne}.
Classifier trained on these constructed samples has good classification ability because the samples are constructed to express characteristic of each attribute of unseen classes. Samples especially after screening are more typical to represent unseen classes and there is a significant increment of classification accuracy by taking screening strategy. Moreover, previous embedding methods inevitably have information loss when associating different spaces but our method can reduce information loss by turning zero-shot learning problem into a supervised learning problem. It's notable that there are no construction methods before to tackle zero-shot learning problem therefore, the result comparison with previous methods is more convictive.
\begin{table}
\small
\centering
  \caption{Comparison to existing approaches in top-1 accuracy (in \%) on four benchmark datasets. The best is marked in bold. $IBSC$: samples without screening; $IBSC^S$: samples after screening.}
  \label{results}
  \begin{tabular}{ccccc}
    \toprule
    Approach & AwA1  & AwA2 & SUN & CUB\\  \hline \hline
    $LATEM$\cite{xian2016latent}& 72.1 & - &-  & 45.6 \\
    $SYNC^{o-vs-o}$\cite{ChangpinyoCGS16} & 69.7 & - &62.8  &53.4\\
    $SYNC^{cs}$\cite{ChangpinyoCGS16} & 68.4& - &52.9  &51.6\\
    $SYNC^{struct}$\cite{ChangpinyoCGS16} & 72.9 & - &62.7 & 54.7\\
    $EXEM(1NN)$\cite{ChangpinyoCS16} & 76.2 & - &69.6 &56.3\\ 
    $EXEM(1NNs)$\cite{ChangpinyoCS16} & 76.5& - &67.3 &\textbf{58.5}\\ \hline
    $IBSC $ & 74.6&  62.4  & 75.5  &36.7\\
    $IBSC^{ s}$& \textbf{82.6} &  \textbf{67.0} & \textbf{80.1}  & 38.4\\
  \bottomrule
\end{tabular}
\end{table}
The relative poor performance on CUB is due to inter-class similarity among unseen classes. There are 200 species of birds in dataset where several species are extremely similar in visual. If source classes have large inter-class similarity, samples constructed from them have weak characteristic expression ability thus, classifier trained on these samples has weak class distinguishing ability. 

\begin{figure}
\includegraphics[height=1.03in, width=2in]{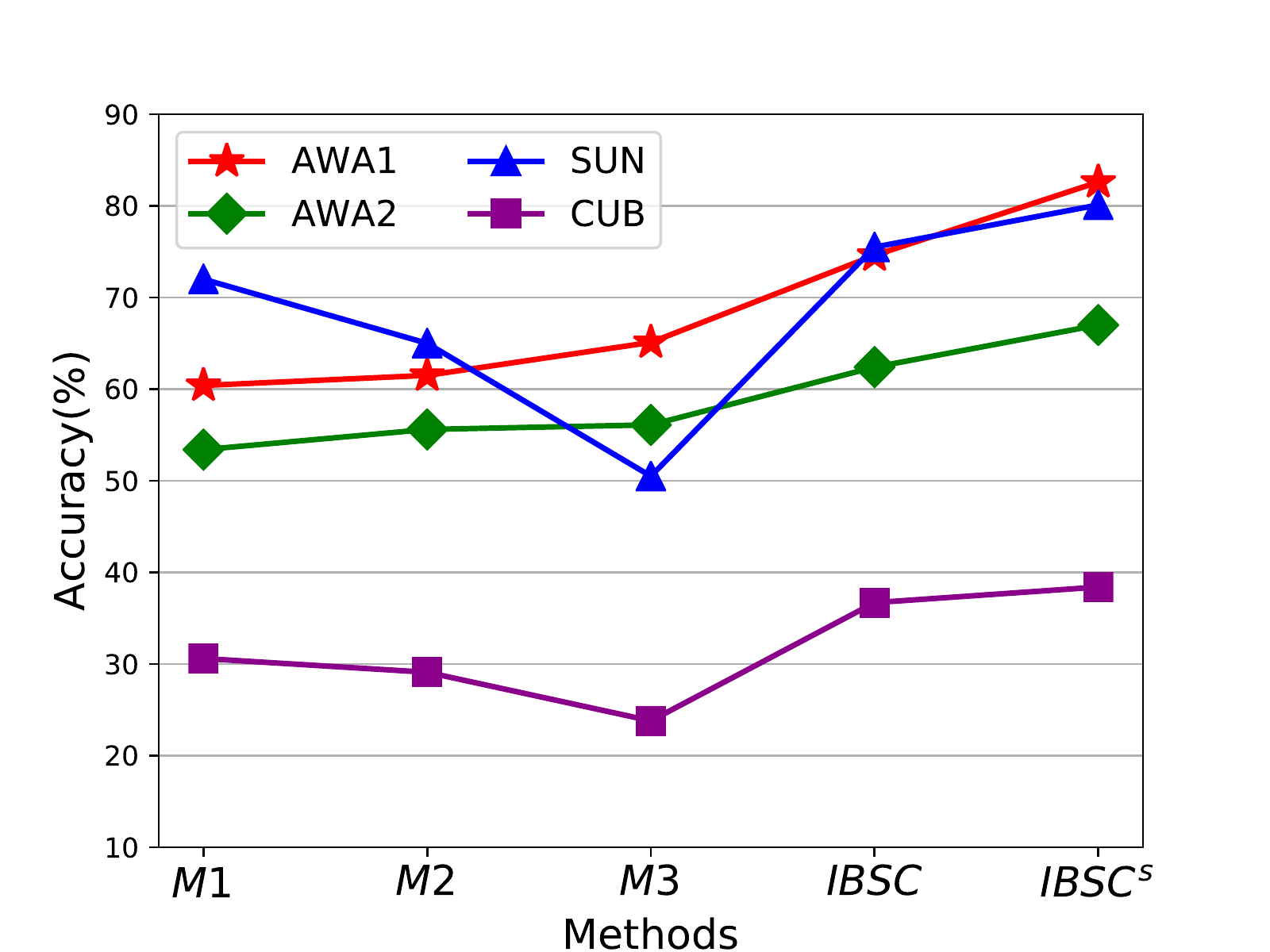}
\caption{Top-1 accuracy of different sample construction methods. Basic ways of constructing samples: $M1$: samples of the most similar classes without changing; $M2$: randomly change value on random feature dimensions of samples in $M1$; $M3$: randomly change value on attribute-relevant feature dimensions of samples in $M1$. $IBSC/IBSC^S$: as illustrated in Table \ref{results}.}
\label{compare}
\end{figure}

We take several basic sample construction strategies to compare with IBSC, as shown in Figure \ref{compare}. M1 uses samples in source classes as training data without any changing while M2 and M3 randomly change feature values. From Figure \ref{compare}, we observe that sample construction based on attribute-feature relation is more reasonable compared with random construction. Although random construction is rough and simple, but in terms of type of methods, it is more comparable than embedding methods.

\section{Conclusions}
We propose a novel method, Imagination Based Sample Construction, to directly construct samples of unseen classes by referring to human associative cognition process. Target samples are constructed by splicing different parts of selected samples of seen classes, which have been proven to have strong capability to characterize unseen classes. The constructed samples are used as labeled data to train a classifier for unseen classes. In this way, we simplify the problem of ZSL into a supervised learning problem. Comprehensive result comparison of four benchmark datasets illustrates the superiority of our method. Moreover, it is the first work concerning sample construction method for ZSL. Therefore, our work can be viewed as a baseline for future sample construction works.

\begin{acks}
This work was supported by the National Natural Science Foundation of China (61773385, 61672523) .

\end{acks}
\bibliographystyle{ACM-Reference-Format}
\bibliography{Imagination}

\end{document}